\documentclass[letterpaper, 10 pt, conference]{ieeeconf}  

\IEEEoverridecommandlockouts                              

\let\labelindent\relax

\font\tenrm = cmr17 at 10pt 

\let\oldnl\nl
\newcommand{\nonl}{\renewcommand{\nl}{\let\nl\oldnl}}

\usepackage{url}

\title{\LARGE \bf
Multi-Robot Informative Path Planning \\ from Regression with Sparse Gaussian Processes
}

\author{Kalvik Jakkala and Srinivas Akella
\thanks{This work was supported in part by NSF under Award Number IIP-1919233. The authors are with the Department of Computer Science, University of North Carolina at Charlotte, Charlotte, NC, USA. Email:{\tt\small \{kjakkala, sakella\}@charlotte.edu}.}%
}

\usepackage{dblfloatfix}
\usepackage{macros}
\begin{document}

\maketitle

\begin{abstract}
This paper addresses multi-robot informative path planning (IPP) for environmental monitoring.  The problem involves determining informative regions in the environment that should be visited by robots to gather the most information about the environment. We propose an efficient sparse Gaussian process-based approach that uses gradient descent to optimize paths in continuous environments. Our approach efficiently scales to both spatially and spatio-temporally correlated environments. Moreover, our approach can simultaneously optimize the informative paths while accounting for routing constraints, such as a distance budget and limits on the robot's velocity and acceleration. Our approach can be used for IPP with both discrete and continuous sensing robots, with point and non-point field-of-view sensing shapes, and for both single and multi-robot IPP. We demonstrate that the proposed approach is fast and accurate on real-world data.
\end{abstract}
\section{Introduction}
Environmental monitoring requires estimating the current state of phenomena, such as temperature, precipitation, ozone concentration, soil chemistry, ocean salinity, and fugitive gas density~(\cite{BinneyKS13, MaLHS18, SuryanT20, JakkalaA22}). Robots can achieve this with informative path planning (IPP). The IPP problem requires strategically determining the regions from which to collect data and the order in which to visit the regions to efficiently and accurately estimate the state of the environment, given finite resources such as the number of robots and battery life.

The IPP problem has been studied in numerous scenarios: persistent ocean monitoring with underwater gliders~\cite{SmithSSJRS11}, information gathering on three-dimensional mesh surfaces for inspection tasks~\cite{ZhuCLSA21}, localizing gas sources in oil fields~\cite{JakkalaA22}, and active learning in aerial semantic mapping~\cite{RuckinJMSP22}.

Most IPP approaches implicitly assume that the environment is correlated~(\cite{SmithSSJRS11, BinneyKS13, HollingerS14, HitzGGPS17, MaLHS18, FrancisOMR19}). Similarly, we consider IPP problems for environments that are correlated either spatially or spatio-temporally and present an efficient approach that leverages such correlations. 

Existing discrete optimization based IPP methods have discretization requirements that limit them to relatively small problems~(\cite{BinneyKS13, MaLHS18}), making them infeasible for large spatio-temporal environments. Additionally, incorporating routing constraints, such as a distance budget and limits on the robot's velocity and acceleration, significantly increase the problem size when using discrete optimization. 

Furthermore, modeling informative paths in continuous domains with potentially continuous sensing robots is a non-trivial problem. The problem is usually addressed using optimization methods such as genetic algorithms and Bayesian optimization~(\cite{HitzGGPS17, FrancisOMR19, VivaldiniMGSORW19, PopovicVHCSSN20}). These methods select sensing locations that maximize mutual information~(MI) computed using Gaussian processes~\cite{krauseSG08}. But these optimization methods are computationally expensive and rely on computing MI, which is also expensive ($\mathcal{O}(n^3)$, where $n$ is the discretized environment size). A few approaches have even considered multi-robot IPP~(\cite{SinghKAGK09, HitzGGPS17}) but they are also inherently limited by the scalability issues of prior IPP approaches. 

Motivated by the above limitations of prior IPP approaches, we present a method that can efficiently generate both discrete and continuous sensing paths, accommodate constraints such as distance budgets and velocity limits, handle point sensors and non-point FoV sensors, and handle both single and multi-robot IPP problems. Our approach leverages gradient descent optimizable sparse Gaussian processes to solve the IPP problem, making it significantly faster compared to prior approaches and scalable to large IPP problems\footnote{Our code is available at \\\url{https://github.com/itskalvik/SGP-Tools}}.
\section{Related Work}
The Informative Path Planning (IPP) Problem is known to be NP-hard~\cite{Brock09TR}. Therefore, only suboptimal solutions can be found for most real-world problems. Numerous IPP methods select utility functions that are submodular~(\cite{ChekuriP05, SinghKAGK09, KrauseG11, BinneyKS13}). Submodular functions have a diminishing returns property that can be leveraged to get good approximation guarantees even when optimized using greedy algorithms. 

Many IPP approaches use mutual information (MI), an information metric that is submodular~\cite{krauseSG08}, as the optimization objective. The methods compute MI using Gaussian processes with known kernel parameters. But MI requires one to discretize the environment, thereby limiting the precision with which the sensing locations can be selected. Also, MI is computationally expensive ($\mathcal{O}(n^3)$, where $n$ is the number of locations in the discretized environment). Singh et al.~\cite{SinghKGK09} proposed a recursive-greedy algorithm that maximized MI. The approach addressed both single and multi-robot IPP. Ma et al.~\cite{MaLHS18} solved the IPP problem by maximizing MI using dynamic programming and used an online variant of sparse Gaussian processes for efficiently learning the model parameters. Bottarelli et al.~\cite{BottarelliBBF19} developed active learning-based IPP algorithms with a complexity of $\mathcal{O}(|D|^5)$, where $D$ is the discretized data collection space.

Hollinger and Sukhatme~\cite{HollingerS14} presented IPP algorithms for continuous spaces that maximized MI using rapidly-exploring random trees~(RRT) and derived asymptotically optimal guarantees. Miller et al.~\cite{MillerSMM16} addressed continuous-space IPP with known utility functions using an ergodic control algorithm. Hitz et al.~\cite{HitzGGPS17} and Popovic et al.~\cite{PopovicVHCSSN20} developed IPP approaches that could optimize the sensing locations in continuous spaces by optimizing any utility function. The approaches used a B-spline to parametrize a path and maximized the utility function (mutual information) using a genetic algorithm. Some authors even leveraged Bayesian optimization~\cite{FrancisOMR19, VivaldiniMGSORW19} to find informative paths in continuous spaces. However, similar to discrete optimization and genetic algorithm based approaches, the method was computationally expensive and limited the approach's scalability. Mishra et al.~\cite{MishraCS18} addressed IPP using dynamic programming, but the approach utilized variance as the information metric, which can be computed quickly but is not as informative as mutual information.

A closely related problem is the correlated orienteering problem (COP), in which one has to plan a path that maximizes the information gain in a correlated environment while restricting the path to a given distance budget~\cite{YuSR14}. Agarwal and Akella~\cite{AgarwalA23} addressed COP for both point locations and 1D features using quadratic programming.

Recently, Rückin et al.~\cite{RuckinJP23} leveraged deep reinforcement learning (DRL) to address the IPP problem. However, it requires significant computational resources to simulate a diverse set of data and train the RL agent.
\section{Multi-Robot Informative Path Planning}
We consider a spatially (or spatiotemporally) correlated stochastic process over an environment $\mathcal{V} \subseteq \mathbb{R}^d$ representing a phenomenon such as temperature. We have $r$ robots and must find the set $\mathcal{P}$ of $r$ paths, one for each robot, so that the data from the phenomenon collected at these locations is sufficient to accurately estimate the phenomenon at every location in the environment. We use the root-mean-square error (RMSE) of the estimates as the measure of accuracy. Since we cannot directly minimize the RMSE, we formulate this problem as one where we want to find the paths $\mathcal{P}$ that maximize the information $I$ (i.e., any function that is a good proxy for accuracy and can be computed without the ground truth labels). Moreover, we also consider constraints $\mathbf{C}$ such as distance budget and velocity limits on the paths:

\begin{equation}
\begin{aligned}
    \mathcal{P}^* = & \argmax_{\{\mathcal{P}_i \in \psi, i = 1,...,r\}} \ I(\cup^r_{i=1} \text{SAMPLE}(\mathcal{P}_i)), \\
    & \text{s.t.} \ \text{Constraints}(\mathcal{P}_{i=1,...,r}) \leq \mathbf{C} \\
\end{aligned}
\end{equation}

Here $\psi$ is the space of paths contained within the environment $\mathcal{V}$, and the $\text{SAMPLE}$ function returns the sensing points along a path $\mathcal{P}_i$. When considering discrete sensing robots, each path is constrained to have only $s$ sensing locations. In a continuous sensing model, the $\text{SAMPLE}$ function returns all the points along the path, which are used to compute the integral of the information collected along the path. In addition, we also consider point sensors such as temperature probes, and non-point sensors that can have any field-of-view (FoV) shape such as a thermal vision camera with a rectangular FoV. 

\section{Preliminaries}

\subsection{Sparse Gaussian Processes}
Gaussian processes (GPs)~\cite{RasmussenW05} are one of the most popular Bayesian approaches. The approach is non-parametric, and its computation cost depends on $n$, the size of the training set. The approach's computation cost is dominated by an expensive $\mathcal{O}(n^3)$ matrix inversion operation on the $n \times n$ covariance matrix, which limits the approach to relatively small datasets that have less than $10,000$  samples. 

The computational cost issues of GPs have been addressed by multiple methods~\cite{SnelsonG06, Titsias09, HoangHL15, BuiYT17}, collectively referred to as sparse Gaussian processes (SGPs). The approaches entail finding a sparse set of $m$ samples called \textit{inducing points} ($m \ll n$), which are used to support the Gaussian process. Since there are fewer samples in the new GPs, these approaches reduce the covariance matrix that needs to be inverted to an $m \times m$ matrix, whose inversion is an $\mathcal{O}(m^3)$ operation. 

The most well-known SGP approach in the Bayesian community is the variational free energy~(VFE) based approach~\cite{Titsias09}. It is a variational approach that is robust to overfitting and is competitive with other SGP approaches. The approach maximizes the evidence lower bound~(ELBO)~$\mathcal{F}$ below to optimize the SGP:

\begin{equation}
\label{vfe}
\begin{aligned}
\mathcal{F} &= \underbrace{\frac{n}{2} \log(2\pi)}_{\text{constant}} + \underbrace{\frac{1}{2} \mathbf{y}^\top (\mathbf{Q}_{nn} + \sigma_{\text{noise}}^2 I)^{-1} \mathbf{y}}_{\text{data fit}} \\
&+ \underbrace{\frac{1}{2} \log |\mathbf{Q}_{nn} + \sigma_{\text{noise}}^2 I|}_{\text{complexity term}} - \underbrace{\frac{1}{2\sigma_{\text{noise}}^2} Tr(\mathbf{K}_{nn} - \mathbf{Q}_{nn})}_{\text{trace term}}\,, \\
\end{aligned}
\end{equation}

where $\mathbf{Q}_{nn} = \mathbf{K}_{nm} \mathbf{K}_{mm}^{-1} \mathbf{K}_{mn}$. The subscripts of the covariance terms $\mathbf{K}$ represent the variables used to compute the matrices; for instance $m$ indicates the inducing points $\mathbf{X}_m$. The training set labels are denoted by $\mathbf{y}$ and $\sigma_{\text{noise}}$ is the data noise variance.

The optimization function $\mathcal{F}$ has three key terms—the data fit, complexity, and trace terms. The data fit term measures prediction accuracy on the training set data. The complexity and trace terms do not depend on the training set labels; instead, they ensure that the SGP-inducing points are well separated and reduce the SGP's overall uncertainty about the training set. When the trace term becomes zero, the SGP becomes equivalent to a full GP. We refer the reader to Bauer et al.~\cite{BauerWR16} for further analysis of the VFE-based SGP~\cite{Titsias09}.

\subsection{SGP-based Sensor Placement}
In our concurrent work~\cite{JakkalaA22b}, we laid the foundation for SGP based sensor placement. We showed that any sensor placement problem can be reduced to a regression problem that can be efficiently solved using sparse Gaussian processes. The method also showed that we can train the SGP in an unsupervised manner by setting the labels of the training set and SGP mean to zero, which disables the label-dependent term of the SGP's optimization function~(Equation~\ref{vfe}). The key advantage of this approach is that it uses the SGP's optimization function as the utility function, which was shown to behave similar to MI while being significantly cheaper to compute than MI. The sensor placement approach, outlined in Algorithm~\ref{alg:Continuous-SGP-short}, entails sampling random unlabeled points in the sensor placement environment and fitting an SGP with known kernel parameters to the sampled points. Once the SGP's inducing points are optimized, they are considered the solution sensor placements.

\begin{algorithm}[ht]
\SetKwInput{kwLoop}{Loop until}
\KwIn{Hyperparameters $\theta$, number of sensor placements~$s$, domain of the environment $\mathcal{V}$}
\KwOut{Sensor placements $\mathcal{A} \in \mathcal{V}$, $|\mathcal{A}| = s$}
$\mathbf{X} \sim \mathcal{U}(\mathcal{V})$ \ \text{/ / \tenrm{Uniformly draw unlabeled locations}} \\ 
$\mathbf{X}_m = \text{RandomSubset}(\mathbf{X}, s) \ \text{/ / \tenrm{Initialize $\mathbf{X}_m$}}$
\text{/ / \tenrm{Initialize SGP with zero mean and unlabeled dataset}} \\
$\varphi = \mathcal{SGP}(\mu=0, \theta, \mathbf{X}, \mathbf{y}=\mathbf{0}, \mathbf{X}_m)$ \\
$\textbf{Loop until} \textit{ convergence}: \mathbf{X}_m \leftarrow \mathbf{X}_m + \gamma \nabla \varphi(\mathbf{X}_m)$ \\
\Return{$\mathbf{X}_m$}
\caption{Continuous-SGP~\cite{JakkalaA22b}: SGP-based sensor placement approach. $\mu$ is the SGP's mean, $\gamma$ is the SGP learning rate.}
\label{alg:Continuous-SGP-short}
\end{algorithm}

\section{SGP-based IPP}

Our SGP based sensor placement approach~\cite{JakkalaA22b} has two key properties that are relevant to addressing the informative path planning (IPP) problem. First, the SGP approach can generate solution sensor placements for both discrete and continuous environments. Second, the approach is able to obtain sensor placement solutions on par with the ones obtained by maximizing mutual information (MI) but with significantly reduced computational cost.

However, the SGP based sensor placement approach does not consider the order in which the sensing locations are visited. Indeed, in IPP, we need to specify the order in which the sensing locations are visited and potentially also consider other constraints on the path, such as distance budget and velocity limits. 

In the following, we first detail our approach to address the visitation order issue of the SGP based sensor placement approach for single-robot IPP. Then we explain how to impose routing constraints, such as a distance budget and velocity limits. After this, we generalize our approach to handle multi-robot IPP, and then finally address continuous sensing along the paths and modeling non-point FoV sensors.

\begin{figure*}[b]
    \centering
    \includegraphics[width=0.75\linewidth]{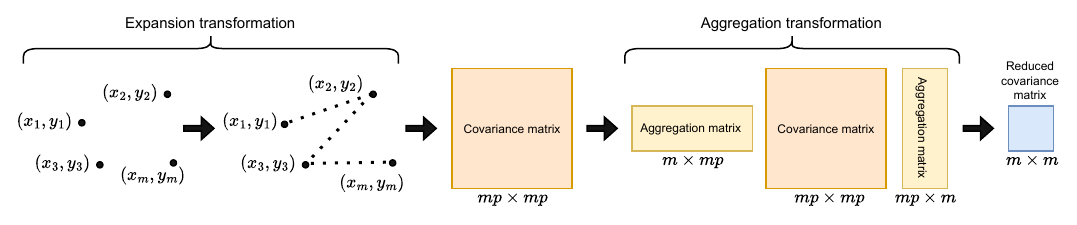}
    \caption{An illustration of the expansion and aggregation transformations used in IPP for continuous sensing robots.}
    \label{fig:transformations}
\end{figure*}

\subsection{Single-Robot IPP}
We address the visitation order issue in spatially correlated environments by leveraging a travelling salesperson problem (TSP) solver~\cite{PerronF19}. In the most fundamental version of the single-robot IPP problem, we need not consider any constraints on the travel distance. Therefore, we first obtain $s$ sensor placement locations using the SGP approach and then generate a path $\cal{P}$ that visits all the solution sensing locations by (approximately) solving the TSP, modified to allow for arbitrary start and end nodes:

\begin{equation}
    \begin{aligned}
        \mathbf{X}_m &= \text{Continuous-SGP}(\theta, \mathcal{V}, s) \\
        \mathbf{X}_m &= \text{TSP}(\mathbf{X}_m) \,.
    \end{aligned}
\end{equation}

In spatio-temporally correlated environments, we do not even have to solve the TSP. This is because the generated solution sensor placements have an inherent visitation order since they span both space and time. However, such an approach could generate solutions that cannot be traversed by real-world robots with restrictions on their dynamics. The approach can handle a distance constraint if allowed to drop a few locations from the selected sensing locations, but it would do that without accounting for the information lost by dropping sensing locations. Therefore, we must develop a more sophisticated approach to address real-world IPP problems that have constraints such as a distance budget and velocity limits.

We do this by leveraging the differentiability of the SGP's optimization objective~$\mathcal{F}$~(Equation~\ref{vfe}) with respect to the inducing points $\mathbf{X}_m$. The inducing points of the SGP $\mathbf{X}_m$, which we consider as the sensing locations, are used to compute the covariance matrix $\mathbf{K}_{mm}$, which is in turn used to compute the Nyström approximation matrix $\mathbf{Q}_{nn}$ in the objective function~$\mathcal{F}$. We can impose constraints on the sensing locations by adding differentiable penalty terms dependent on the inducing points $\mathbf{X}_m$ to the objective function~$\mathcal{F}$. Such an objective would still be differentiable and can be optimized using gradient descent.

We use the above method to impose constraints on even the solution \textit{paths}. We do this by first solving the TSP on the SGP's initial inducing points $\mathbf{X}_m$ and treating them as an ordered set, which gives us an initial path that sequentially visits the inducing points. We then augment the SGP's objective function~$\mathcal{F}$ with differentiable penalty terms that operate on the ordered inducing points for each path constraint and optimize the SGP to get the solution path. For instance, we can formulate the distance budget constraint as follows:

\begin{equation}
\begin{aligned}
\label{eq:distance_constraint}
\hat{\mathcal{F}} =& \ \mathcal{F} - \alpha \text{ReLu}(\text{PathLength}(\mathbf{X}_m)-c)  \,, \\
&\text{where ReLu}(x) = \text{max}(x, 0) \,.
\end{aligned}
\end{equation}

Here, PathLength is a function to obtain the total travel distance of the path that sequentially visits each of the inducing points $\mathbf{X}_m$ (treated as an ordered set) and $\alpha$ is a weight term used to scale the distance constraint penalty term. The ReLu function ensures that $\mathcal{F}$ remains unchanged if the path length is within the distance budget $c$ and penalizes it only if the length exceeds the distance budget. 

Similarly, we can accommodate additional constraints on the route, such as limits on the velocity and acceleration. In addition, we can trivially set predefined start and end points for the paths by freezing the gradient updates to the first and last inducing points of the SGP.

\subsection{Multi-Robot IPP}
\label{sec:multi}

We now address the multi-robot IPP problem. This is achieved by increasing the number of inducing points in the SGP. If we have $r$ robots and need paths with $s$ sampling locations each, we sample $rs$ random points from the environment and find $r$ paths by solving the vehicle routing problem~(VRP, \cite{TothV14}). After resampling the paths so that each path has $s$ points, this yields an ordered set with $rs$ sensing locations that form the $r$ initial paths. We initialize the SGP with these $rs$ points as the inducing points. We then add the path constraints that operate on each path to the objective function~$\mathcal{F}$ and optimize the SGP to get the set of $r$ solution paths $\mathcal{P}$. The approach is shown in Algorithm~\ref{alg:Multi-IPP}. For spatio-temporal environments, we can also decouple the spatial and temporal inducing points to further reduce the computational cost of our approach and ensure that it has optimal waypoint assignments (see the Appendix~\cite{JakkalaA24}).

\begin{algorithm}[ht]
\SetKwInput{kwLoop}{Loop until}
\KwIn{Hyperparameters $\theta$, domain of the environment $\mathcal{V}$, number of waypoints $s$, number of robots $r$, path constraints $\mathbf{C}$}
\KwOut{$\mathcal{P} = \{\mathcal{P}_i | \mathcal{P}_i \in \mathcal{V}, |\mathcal{P}_i| = s, i=1,...,r\}$}
$\mathbf{X} \sim \mathcal{U}(\mathcal{V})$ \ \text{/ / \tenrm{Uniformly draw unlabeled locations}} \\ 
$\mathbf{X}_m = \text{RandomSubset}(\mathbf{X}, rs)$ \ \text{/ / \tenrm{Initialize $\mathbf{X}_m$}} \\
$\mathbf{X}_m = \text{VRP}(\mathbf{X}_m)$ \ \text{/ / \tenrm{Get set of initial paths $\mathcal{P}$}} \\
\text{/ / \tenrm{Add constraints to the SGP's optimization function $\mathcal{F}$}} \\
$\hat{\mathcal{F}} = \ \mathcal{F} - \alpha (\text{Constraints}(\mathbf{X}_m)-\mathbf{C})$ \\
\text{/ / \tenrm{Initialize SGP with the ordered inducing points}} \\
$\varphi = \mathcal{SGP}(\mu=0, \theta, \mathbf{X}, \mathbf{y}=\mathbf{0}, \mathbf{X}_m, \hat{\mathcal{F}})$ \\
$\textbf{Loop until} \textit{ convergence}: \mathbf{X}_m \leftarrow \mathbf{X}_m + \gamma \nabla \varphi(\mathbf{X}_m)$ \\
\Return{$\mathbf{X}_m$}
\caption{SGP-IPP: Multi-Robot IPP approach. $\mu$ is the SGP's mean, $\gamma$ is the SGP learning rate, and VRP is the vehicle routing problem solver.}
\label{alg:Multi-IPP}
\end{algorithm}

\subsection{IPP for Continuous and Non-point FoV Sensing Robots} 

Our approach so far has only considered discrete sensing robots with point sensors. However, there are instances where we require robots to continuously sense along their paths, or to utilize sensors such as cameras with non-point fields of view (FoV). While continuous and non-point FoV sensing robots can use paths optimized for discrete sensing robots with point sensors, explicitly optimizing paths for continuous and non-point FoV sensing robots is likely to yield more informative paths.

A naive approach to modeling continuous sensing robots is to use a large number of inducing points. However, such an approach would limit scalability due to the cubic complexity of SGPs with respect to the number of inducing points. Additionally, it cannot handle non-point FoV sensors, as even if the additional inducing points are initialized to match the sensor's FoV, they would not retain the FoV shape after optimization. A key advantage of generalizing the SGP-based sensor placement approach~\cite{JakkalaA22b} to IPP is that we can leverage the properties of GPs and SGPs~\cite{RasmussenW05, SmithP05, SnelsonG06, Titsias09}. We describe two such properties and how they can be used to address IPP for continuous and non-point FoV sensing robots.

First, we utilize the property that the inducing points of SGPs can be transformed with any non-linear function and still be optimized using gradient descent. We can employ such transformations to approximate the information along solution paths. To achieve this, we parameterize the $m$ inducing points of the SGP as the sensing locations for a discrete point sensing robot's path. Then, we apply a transformation—the expansion transformation $T_\text{exp}$—to interpolate $p$ additional points between every consecutive pair of inducing points that form the robot's path $\mathbf{X}_{mp} = T_\text{exp}(\mathbf{X}_{m})$, resulting in $mp$ inducing points (actually $(m-1)p$ points, denoted as $mp$ to simplify notation). We then utilize the $mp$ inducing points $\mathbf{X}_{mp}$ to compute the SGP's objective function $\hat{\mathcal{F}}$ with path constraints. Note that the interpolation operation is differentiable, enabling us to compute gradients for the $m$ inducing points $\mathbf{X}_{m}$. This approach enables us to consider the information gathered along the entire path.

Next, we leverage the property of GPs that they are closed under linear transformations~\cite{RasmussenW05, SmithSSJRS11}. We use the aggregation transformation $T_\text{agg}$, which aggregates (via averaging) the SGP's covariances corresponding to the $p$ inducing points that approximate the information between every consecutive pair of waypoints of the path (i.e., the $m$ inducing points), thereby reducing the size of the covariance matrix from $mp \times mp$ back to $m \times m$. Specifically, we first employ the expansion transformation $T_\text{exp}$ on the $m$ inducing points to map them to a larger set of $mp$ points. Then, we apply the aggregation transformation $T_\text{agg}$ to the covariance matrices built using the $mp$ points. These covariances are utilized to compute $\mathbf{Q}_{nn}$, which in turn is used to compute the SGP's objective function (Equation~\ref{vfe}):

\vspace{-3mm}
\begin{equation}
\begin{aligned}
\label{Q_nn}
\mathbf{Q}_{nn} &= \mathbf{K}_{n \times mp}T_\text{agg} (T_\text{agg}^\top \mathbf{K}_{mp \times mp}T_\text{agg})^{-1} T_\text{agg}^\top \mathbf{K}_{mp \times n}\,.
\end{aligned}
\end{equation}

Here, $\mathbf{K}_{n \times mp}$ represents the covariance matrix between the $n$ training set inputs and the $mp$ inducing points. The aggregation transformation reduces the size of the covariance matrix $\mathbf{K}_{mp \times mp}$ before inversion. Consequently, the matrix inversion cost is reduced to $\mathcal{O}(m^3)$ from $\mathcal{O}(m^3p^3)$, allowing us to benefit from both the expansion transformation, which enables modeling of continuous sensing robots, and the reduced computational cost from the aggregation transformation. Note that the information along the path between the waypoints is retained in the aggregated covariances. Additionally, we observed that the aggregation transformation stabilized the gradients during the optimization of the inducing points. The approach is illustrated in Figure~\ref{fig:transformations}; see further details of the transformations in the Appendix \cite{JakkalaA24}.

We can use the transformations detailed above to handle non-point field of view (FoV) sensing robots as well. This is accomplished by employing the expansion transformation to map each of the $m$ inducing points to $p$ points that approximate the sensor's FoV area. Additionally, since the gradients are propagated back to the original $m$ inducing points $\mathbf{X}_{m}$, the method retains the FoV shapes. Furthermore, we can leverage this property to model sensors with integrated observations such as gas sensors~\cite{JakkalaA22, LongiRSMRSHK20}, where the labels are modeled as $y_i = ||\mathbf{w}_i|| \int_0^1 f(\mathbf{w}_i t+\mathbf{z}_i)dt + \epsilon_i$, with $\mathbf{z}_i$ as the start point of a line segment along which the data is integrated, and $\mathbf{w}_i$ giving the direction and length of the line segment. Similarly, we can efficiently model complex path parametrizations, such as using splines to obtain smooth paths, account for sensors such as cameras whose FoV varies with height from the ground, and even model FoVs that consider the shape of the surface, such as stereo vision cameras when used to scan 3D surfaces. For further details, please refer to the Appendix~\cite{JakkalaA24}.
\section{Experiments}

\begin{figure*}
   \centering
   \subfloat[RMSE (ROMS) \label{fig:benchmark-rmse-sal}]{
    \includegraphics[width=0.24\linewidth]{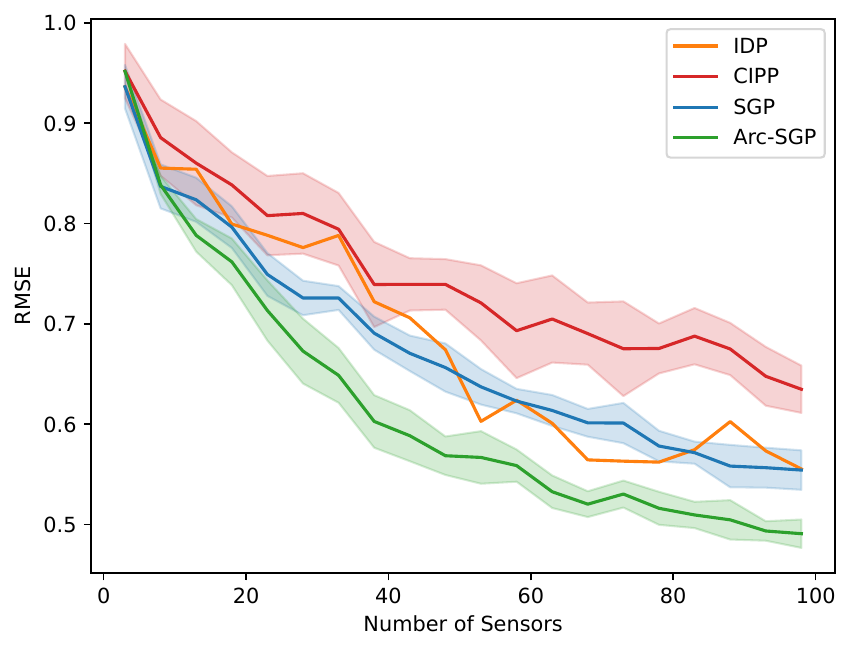}
    }
   \subfloat[RMSE (Soil) \label{fig:benchmark-rmse-soil}]{
    \includegraphics[width=0.24\linewidth]{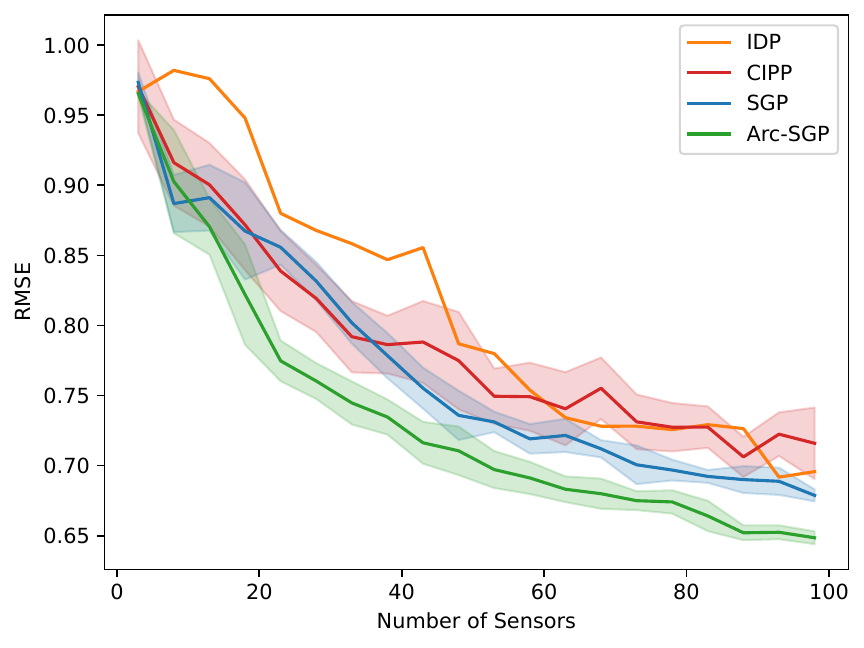}
    }
   \subfloat[Runtime (ROMS) \label{fig:benchmark-time-sal}]{
    \includegraphics[width=0.24\linewidth]{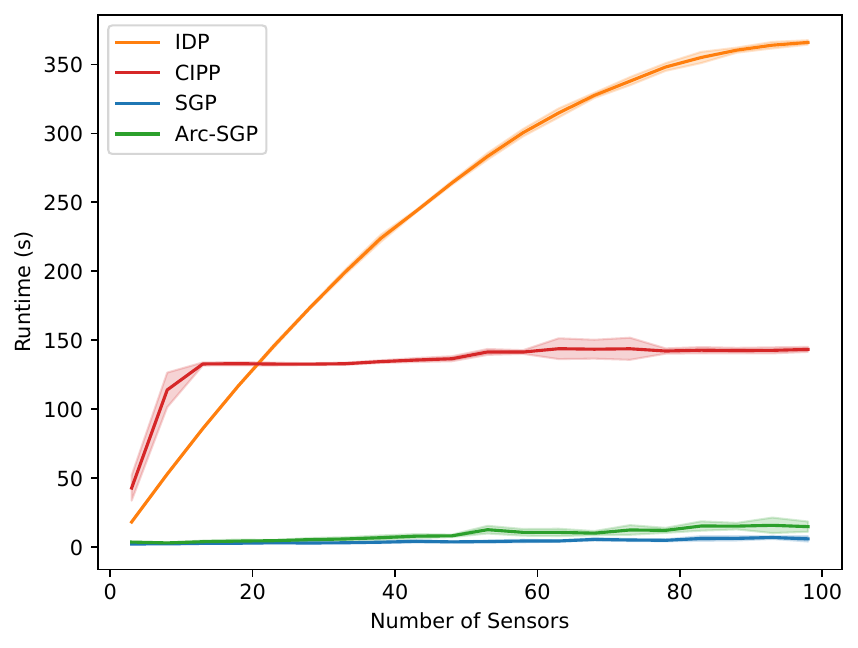}
    }
   \subfloat[Runtime (Soil) \label{fig:benchmark-time-soil}]{
    \includegraphics[width=0.24\linewidth]{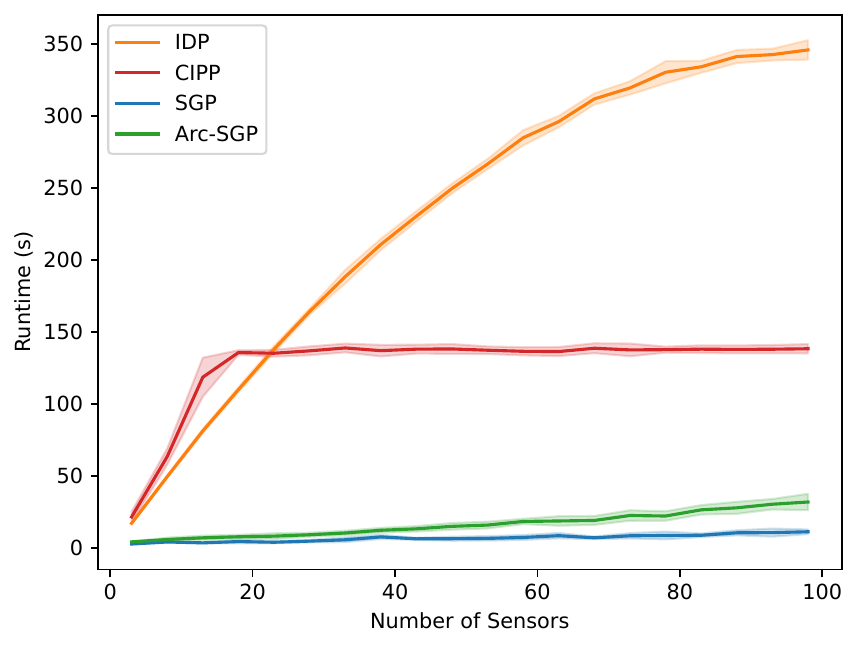}
    }
  \caption{RMSE and runtime results for single robot IPP with the IDP, CIPP, SGP, and Arc-SGP approaches on the ROMS and US soil datasets.}
  \label{fig:benchmark-1R}
\end{figure*}

 \begin{figure*}
   \centering
   \subfloat[RMSE (ROMS) \label{fig:benchmark-rmse-sal-nR}]{
    \includegraphics[width=0.24\linewidth]{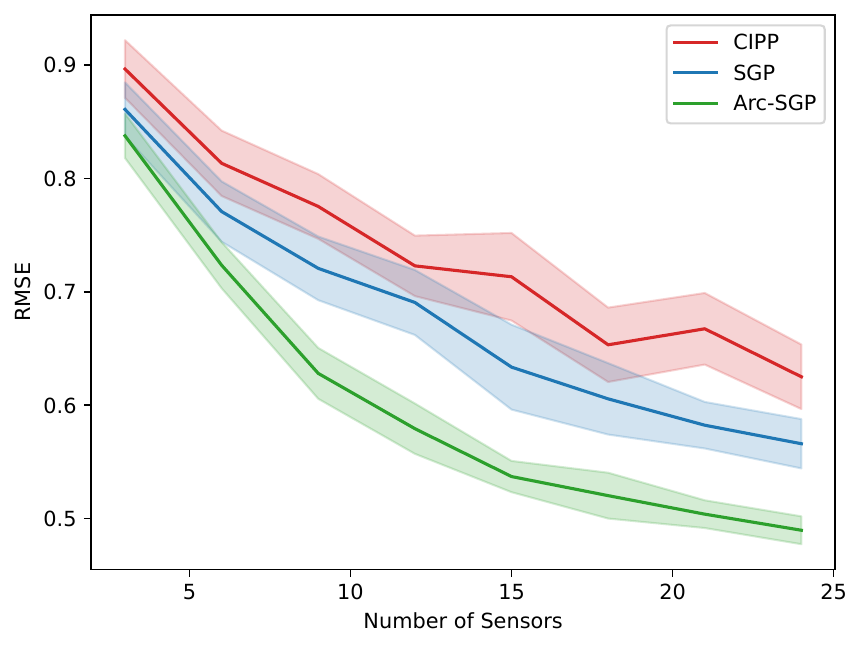}
    }
   \subfloat[RMSE (Soil) \label{fig:benchmark-rmse-soil-nR}]{
    \includegraphics[width=0.24\linewidth]{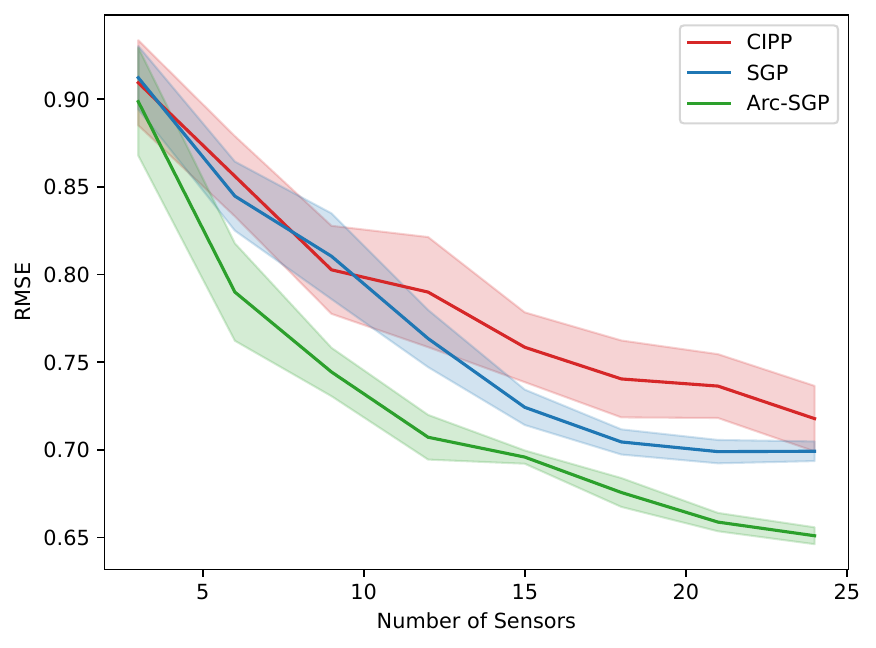}
    }
   \subfloat[Runtime (ROMS) \label{fig:benchmark-time-sal-nR}]{
    \includegraphics[width=0.24\linewidth]{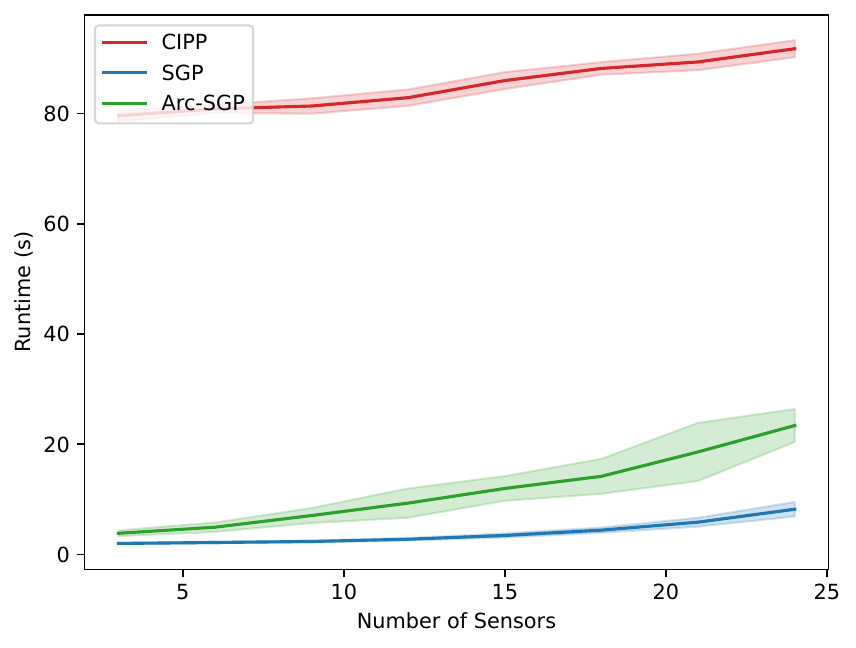}
    }
   \subfloat[Runtime (Soil) \label{fig:benchmark-time-soil-nR}]{
    \includegraphics[width=0.24\linewidth]{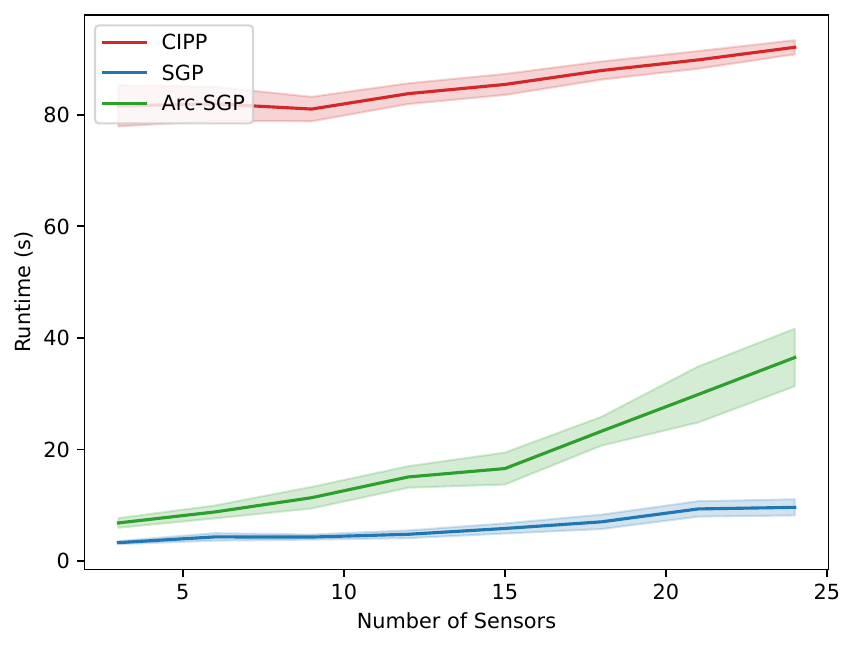}
    }
  \caption{RMSE and runtime results for four-robot IPP with the CIPP, SGP, and Arc-SGP approaches on the ROMS and US soil datasets.}
  \label{fig:benchmark-nR}
\vspace{-3mm}
\end{figure*}

We first demonstrate our approach for the unconstrained single robot IPP problem on the ROMS ocean salinity~\cite{Shchepetkin05} and US soil moisture~\cite{SoilData} datasets. The ROMS dataset contains salinity data from the Southern California Bight region, and the US soil moisture dataset contains moisture readings from the continental USA. All experiments were benchmarked on a workstation with an 18-core Intel(R) Xeon(R) Gold 6154 3.00GHz CPU and 128 GB of RAM. 

We benchmarked our SGP based IPP approach (SGP) that optimizes a path for discretely sensing $s$ locations and our transformation based generalization of the SGP based IPP approach (Arc-SGP) that optimizes the paths while accounting for the information collected along the entire path. We also benchmarked two baseline approaches—the Information-Driven Planner (IDP, Ma et al.~\cite{MaLHS18}) and Continuous-Space Informative Path Planner (CIPP, Hitz et al.~\cite{HitzGGPS17}). IDP leverages discrete optimization to iteratively find discrete sensing locations that maximize mutual information (MI), and CIPP leverages CMA-ES, a genetic algorithm, to find informative sensing locations that maximize MI in continuous spaces.

An RBF kernel~\cite{RasmussenW05} was used to model the spatial correlations of the datasets (the baselines use it to measure MI). We evaluated the paths by gathering the ground truth data along the entire generated solution paths (i.e., by continuous sensing robots) and estimating the state of the whole environment from the collected data. The root-mean-squared error (RMSE) between the ground truth data and our estimates was used to quantify the quality of solution paths. We generated solution paths for both the datasets with the number of path sensing locations ranging from 3 to 100 in increments of 5. The experiment was repeated 10 times. The mean and standard deviation of the RMSE and runtime results on the ROMS and US soil moisture datasets are shown in Figure~\ref{fig:benchmark-1R}.

As we can see, our SGP approach is consistently on par or better than the two baselines in terms of RMSE, and our Arc-SGP approach has a considerably lower RMSE than the other approaches in all cases. Also, both our approaches substantially outperform the baselines in computation time (up to 35 times faster). In both the baselines, a significant amount of computation time is spent on computing MI, while our SGP approach's objective approximates the same in a computationally efficient manner (detailed in our foundational work~\cite{JakkalaA22b}). Indeed, the MI computation cost is the key reason why both IDP and CIPP cannot scale to spatio-temporally correlated environments, since even with a coarse discretization, it would be far too computationally expensive. Also, since our approaches rely on gradient information, they are significantly faster to converge compared to the discrete and genetic algorithm based baseline approaches. 

We now demonstrate our approach for multi-robot IPP. We used the same kernel parameters as we did in the previous experiments. The solution paths were generated for four robots with the number of optimization waypoints ranging from 3 to 25 in increments of 5 for each robot's path. We evaluated the SGP, Arc-SGP, and CIPP methods, which support multi-robot IPP. The RMSE and runtime results are shown in Figure~\ref{fig:benchmark-nR}. Our SGP approach consistently performs on par with or better than the CIPP approach in terms of RMSE. Additionally, Arc-SGP achieves notably lower RMSE compared to both SGP and CIPP, as it explicitly optimizes the paths for continuous sensing. Moreover, both our approaches—SGP and Arc-SGP—significantly outperform CIPP (up to 26 times faster) in terms of compute time.

\begin{figure}
    \centering
    \includegraphics[width=0.3\linewidth]{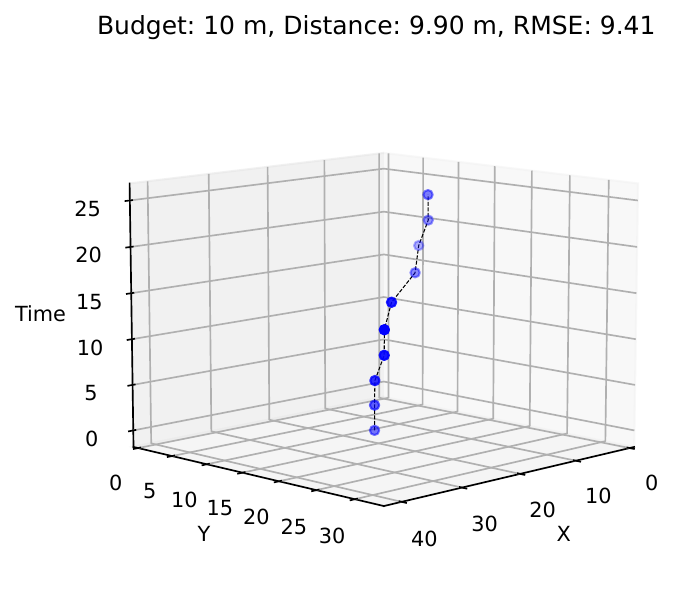}
    \hfill
     \includegraphics[width=0.3\linewidth]{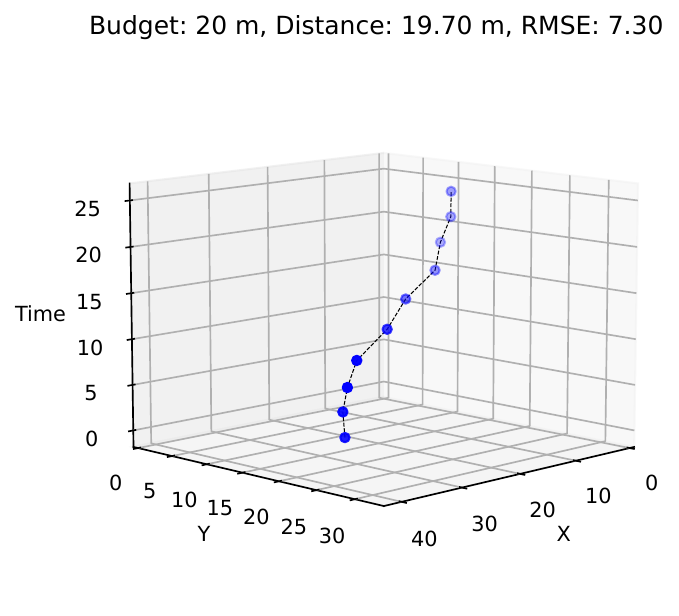}
    \hfill
     \includegraphics[width=0.3\linewidth]{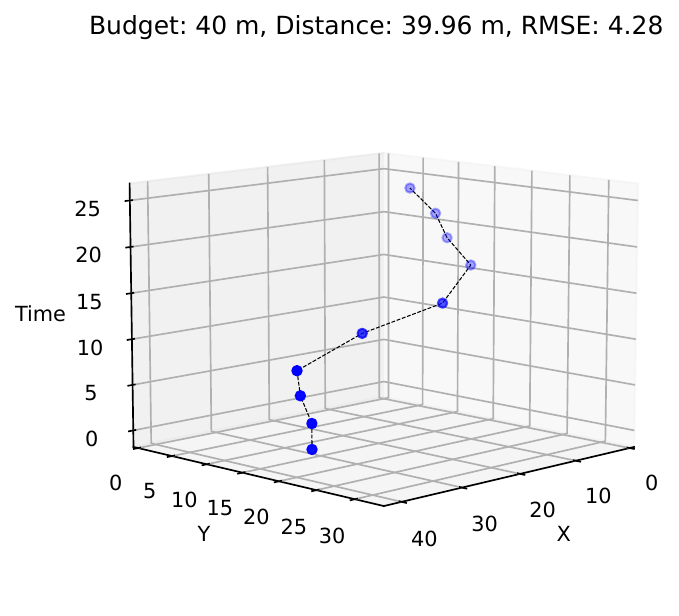}
    \hfill
	\caption{Data collection paths generated using a spatio-temporal kernel function for different distance budgets.}
  \label{fig:dist}
\vspace{-5mm}
\end{figure}

\begin{figure}
    \centering
    \includegraphics[width=0.3\linewidth]{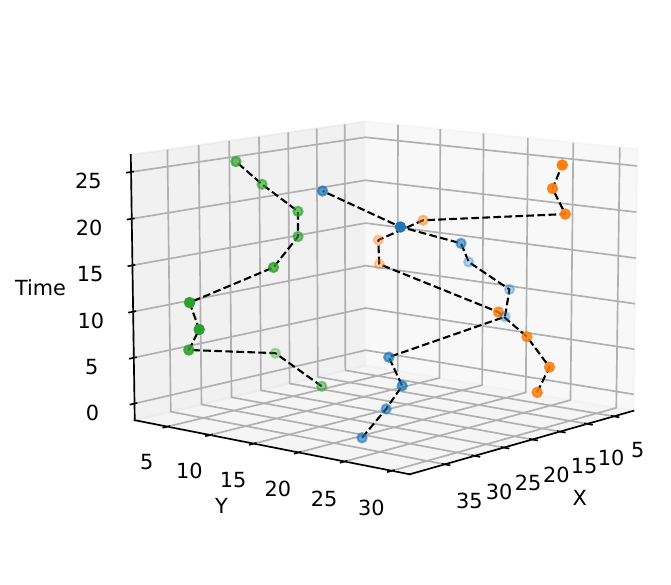}
    \hfill
    \includegraphics[width=0.3\linewidth]{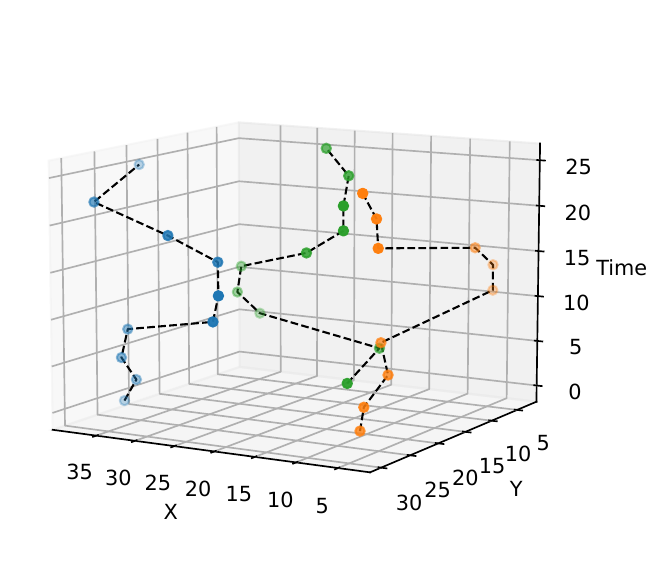}
    \hfill
    \includegraphics[width=0.3\linewidth]{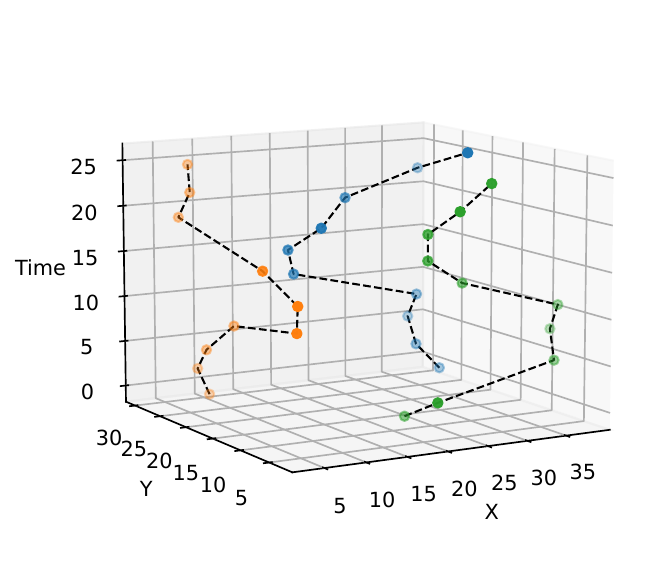}
    \hfill
	\caption{Three different views of our multi-robot IPP solution paths for a spatio-temporal data field, with path lengths of 47.29 m, 47.44 m, and 47.20 m. The total RMSE was 2.75.}
  \label{fig:multi}
\vspace{-3mm}
\end{figure}

We next show our approach for spatio-temporal IPP with a distance constraint. A Gaussian process was used to sample dense spatio-temporal temperature data. We used an RBF kernel with a length scale of 7.70 m, 19.46 m, and 50.63 mins along the $x$, $y$, and temporal dimensions, respectively. We generated paths by optimizing the inducing points in our SGP approach with distance budgets of 10 m, 20 m, and 40 m; the results are shown in Figure~\ref{fig:dist}. Our approach consistently saturates the distance budget without exceeding it to get the maximum amount of new data, evident from the paths' RMSE scores. We also show the paths generated for three robots in the same environment (Figure~\ref{fig:multi}). We do not show the reconstructions since the data is spatio-temporal, which is difficult to show in 2D.

\begin{figure}
    \centering
    \includegraphics[width=0.45\linewidth]{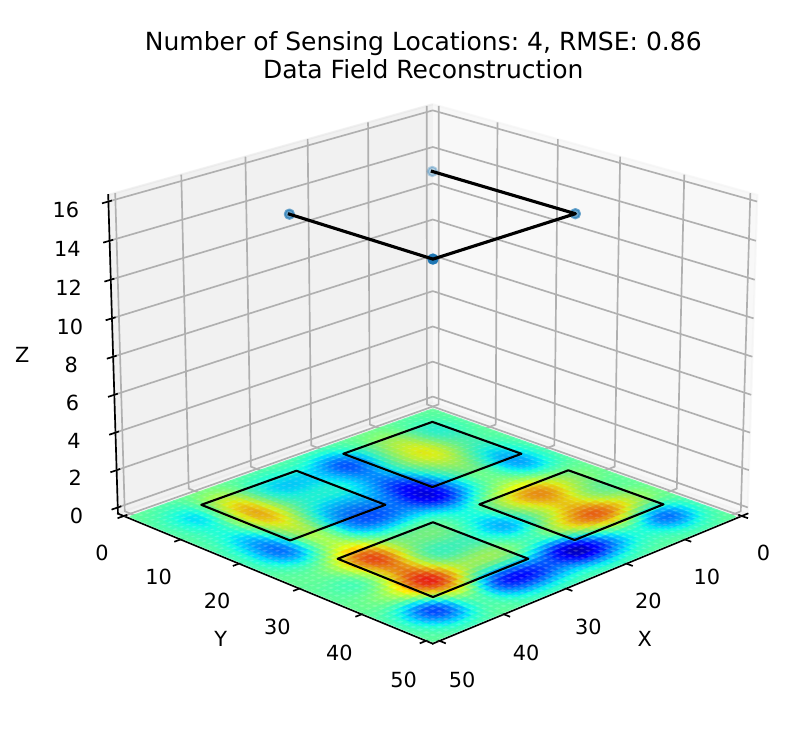}
    \hfill
    \includegraphics[width=0.45\linewidth]{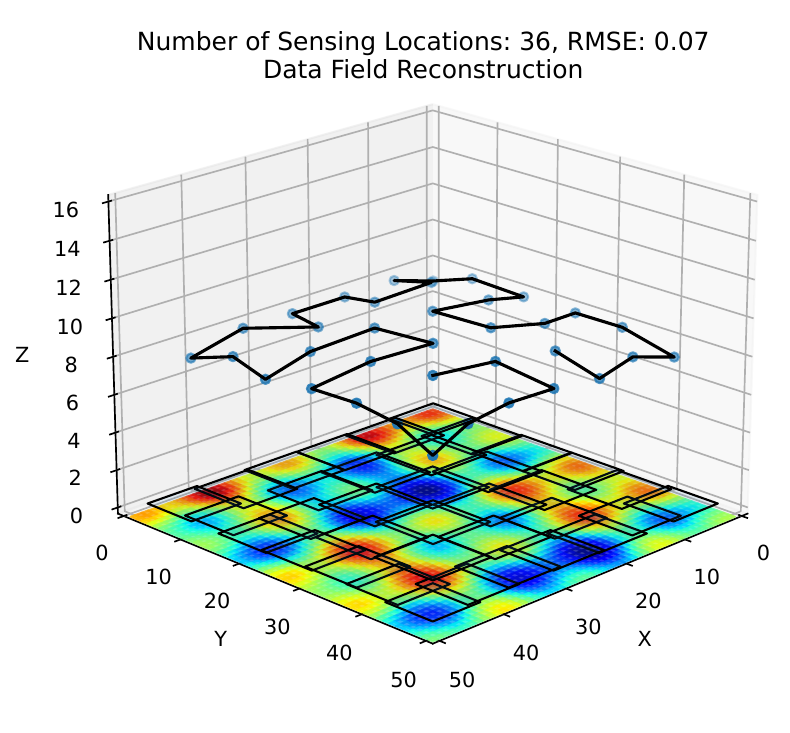}
	\caption{Solution paths for a discrete sensing robot with a square height-dependent FoV area sensor: (left) 4 sensing locations, (right) 36 sensing locations. The sensing height is optimized to ensure a good balance between the ground sampling resolution and the FoV area (black squares).}
  \label{fig:height}
\vspace{-3mm}
\end{figure}

Figure~\ref{fig:height} shows our SGP approach for a discrete sensing robot, i.e., it senses only at the path's vertices (blue points). We considered a 3D environment with densely sampled elevation data and parametrized the path so that we account for the robot's sensing FoV area to be a function of the robot's height from the ground. We used an RBF kernel with a length scale of 3 m (details in the Appendix~\cite{JakkalaA24}). 

Also, note that our approach can solve for paths in more complex environments and utilize non-stationary kernels~\cite{RasmussenW05} to capture intricate correlation patterns within environments. Furthermore, while this paper primarily focuses on path planning, we can extend our approach to encompass full trajectory planning. This involves optimizing the paths to accommodate the velocity, acceleration, and turning radius constraints of the robots, along with any other dynamics constraints, by parameterizing the path as a B-spline. 
\section{Conclusion}
We presented an efficient continuous space approach to informative path planning using sparse Gaussian processes that can address various challenges related to monitoring in spatially and spatio-temporally correlated environments. Our approach can model routing constraints, and handle discrete and continuous sensing robots with arbitrary FoV shapes. Furthermore, our method generalizes to multi-robot IPP problems as well. We demonstrated that our approach is fast and accurate for IPP on real-world data. We also presented our IPP solutions for different distance budgets, multi-robot scenarios, and with non-point FoV sensing robots. Our future work will build upon this approach to extend its applicability to online and decentralized IPP problems.


\bibliography{references.bib}

\end{document}


\maketitle

\tableofcontents

\section{Additional Method Details}

\subsection{Aggregation Transformation}
We found that the aggregation transformation, in addition to reducing computation costs, also stabilized gradients while optimizing the inducing points. This is because without the aggregation transformation, gradients from the objective function are directly propagated to each of the $p$ inducing points after the expansion transformation, which approximate the Field of View (FoV) of a sensor. Consequently, the gradients of the points within the FoV of a sensor could all be directed away from each other, resulting in a net zero gradient being propagated back to the original inducing point from before the expansion transformation. However, when using the aggregation transformation, the gradients are first computed with respect to the aggregated covariances of all the $p$ inducing points approximating the FoV of a sensor. Then, the gradients are further propagated back to the original inducing point from before the expansion transformation through the aggregation transformation operation. This results in the net gradients to the original $m$ inducing points better representing the gradients for moving the whole FoV instead of the individual points within the FoV of each sensor.

\subsection{Modeling Complex Path Parameterizations}
The expansion transformation can be used to efficiently model complex path parametrizations, such as using splines to create smooth paths, accommodating sensors like cameras whose Field of View (FoV) varies with height from the ground, and even modeling FoVs that consider the shape of the surface, such as stereo vision cameras used for scanning 3D surfaces.

Consider a sensor like a camera whose FoV varies with height from the ground. We can model such a sensor by parameterizing the $m$ inducing points with three dimensions: two for the x and y axes, respectively, and one for the height of the camera from the ground. The expansion transformation can then be used to map each $m$ 3D-inducing points to a set of 2D-inducing points in the xy ground plane, approximating the area covered by the sensor. Since we have access to the height of the camera from the ground, we can use that information to determine the spacing between the rows and columns of the 2D grid of points approximating the ground coverage of each sensor. Please refer to our code repository for the code used to generate our results.

\clearpage
\section{Precipitation Dataset Experiments}
\begin{figure}[ht!]
   \centering
   \subfloat[RMSE results.\label{fig:benchmark-rmse-pre}]{
    \includegraphics[width=0.37\linewidth]{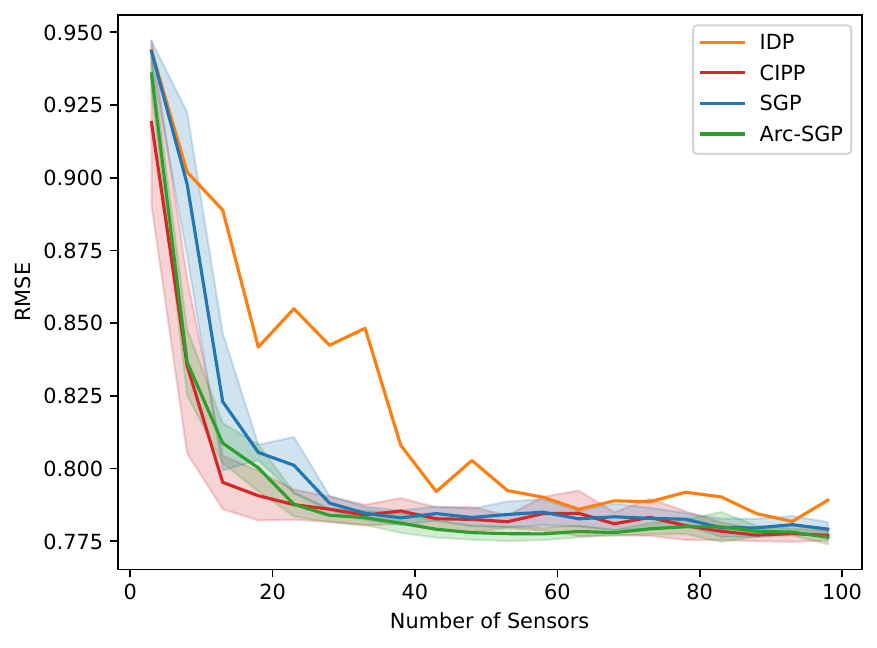}
    }
   \subfloat[Runtime results.\label{fig:benchmark-time-soil}]{
    \includegraphics[width=0.37\linewidth]{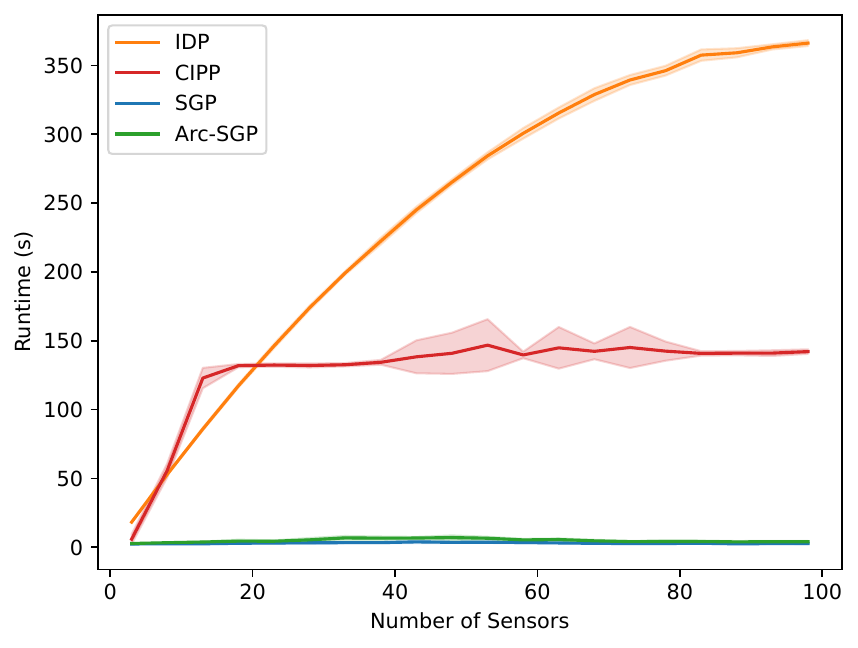}
    }
  \caption{RMSE and runtime results of the IDP, SGP, Arc-SGP, and CIPP approaches on the precipitation dataset.}
  \label{fig:benchmark-pre}
\end{figure}

\begin{figure}[ht!]
   \centering
   \subfloat[RMSE results.\label{fig:benchmark-rmse-sal-nR}]{
    \includegraphics[width=0.37\linewidth]{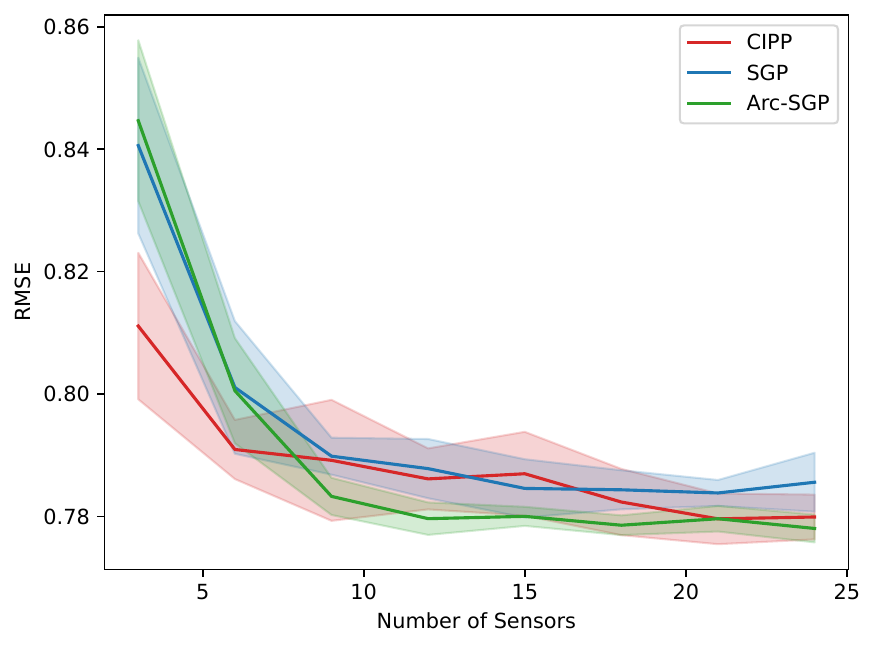}
    }
   \subfloat[Runtime results.\label{fig:benchmark-time-sal-nR}]{
    \includegraphics[width=0.37\linewidth]{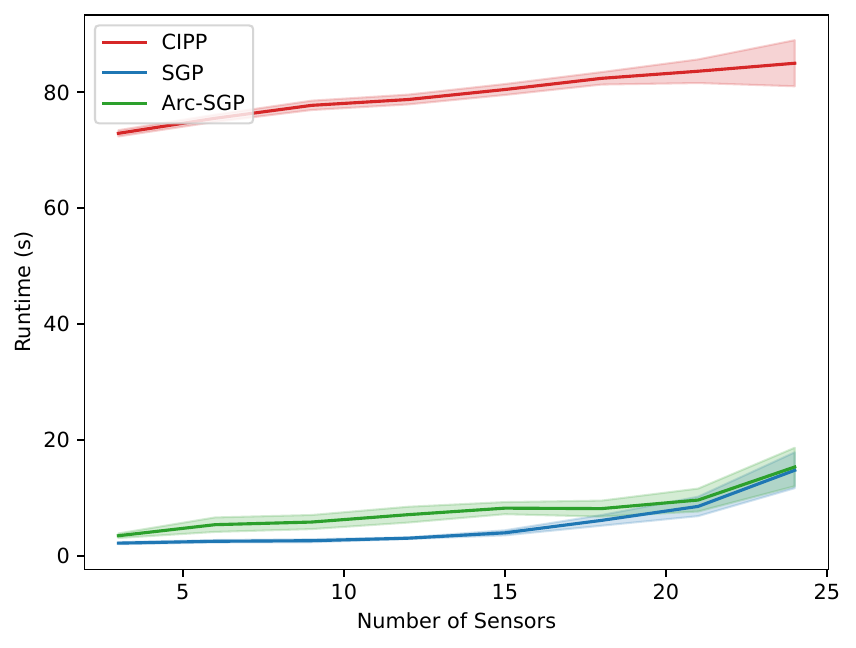}
    }
  \caption{RMSE and runtime results of the SGP, Arc-SGP, and CIPP approaches on the precipitation dataset.}
  \label{fig:benchmark-pre-nR}
\end{figure}

This section presents our results on the precipitation~\cite{brethertonWDWB99} dataset. The precipitation dataset contains daily precipitation data from 167 sensors around Oregon, U.S.A., in 1994. The experimental protocols were the same as the benchmarks presented in the main paper on the other two datasets. Figure~\ref{fig:benchmark-pre} and Figure~\ref{fig:benchmark-pre-nR} show the single robot and multi-robot IPP results, respectively. We see that the RMSE results saturate to similar scores with fewer number of sensors, but still closely follow the trends of the benchmark results shown in the main paper. In other words, our SGP-based approaches are consistently on par or better than the baselines in terms of RMSE. Additionally, our approaches' computation time is significantly lower than the baselines.

\section{Figures}

\begin{figure}[b]
    \centering
    \includegraphics[width=\linewidth]{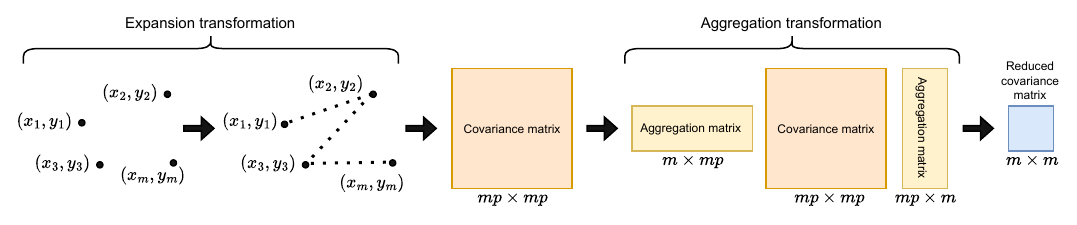}
    \caption{An illustration of the expansion and aggregation transformations used in IPP for continuous sensing robots.}
    \label{fig:transformations}
\end{figure}

\begin{figure}
    \centering
    \includegraphics[width=0.3\linewidth]{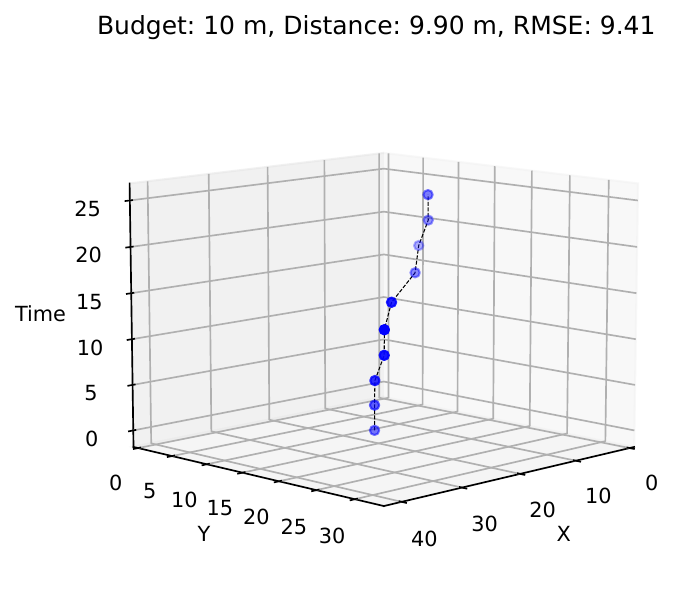}
    \hfill
     \includegraphics[width=0.3\linewidth]{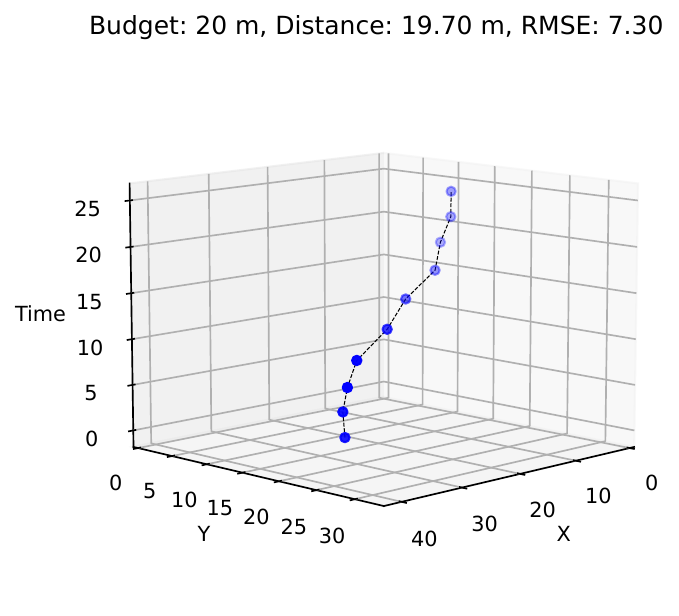}
    \hfill
     \includegraphics[width=0.3\linewidth]{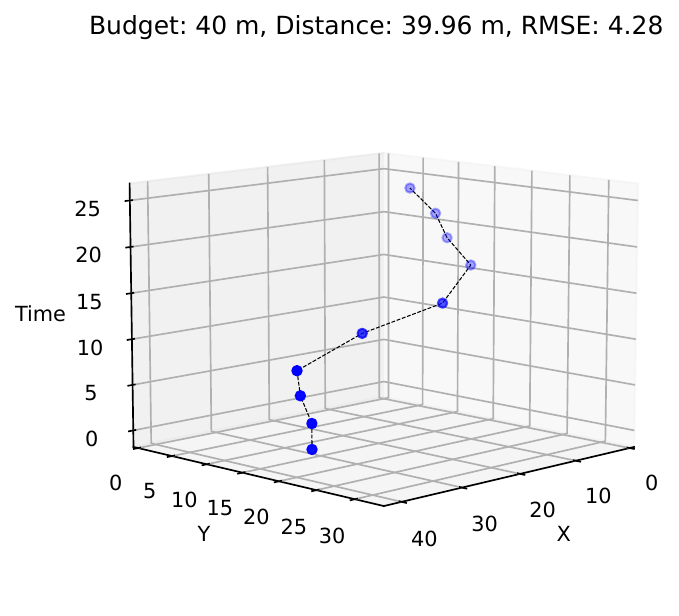}
    \hfill
	\caption{Data collection paths generated using a spatio-temporal kernel function for different distance budgets.}
  \label{fig:dist}
\end{figure}

\begin{figure}
    \centering
    \includegraphics[width=0.3\linewidth]{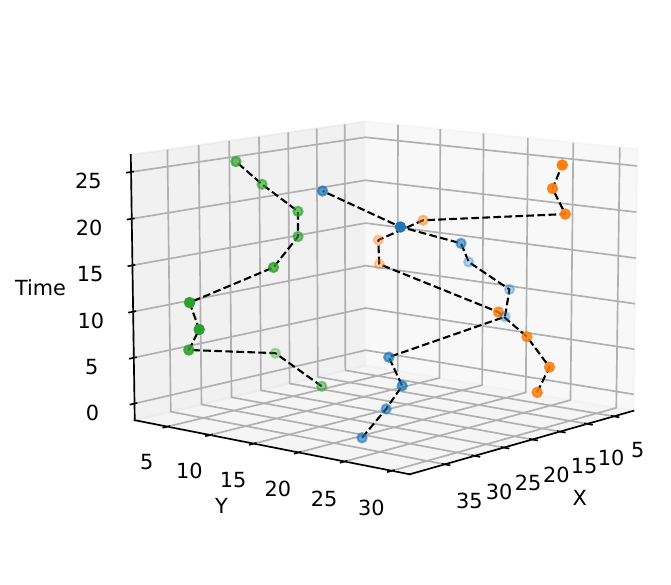}
    \hfill
    \includegraphics[width=0.3\linewidth]{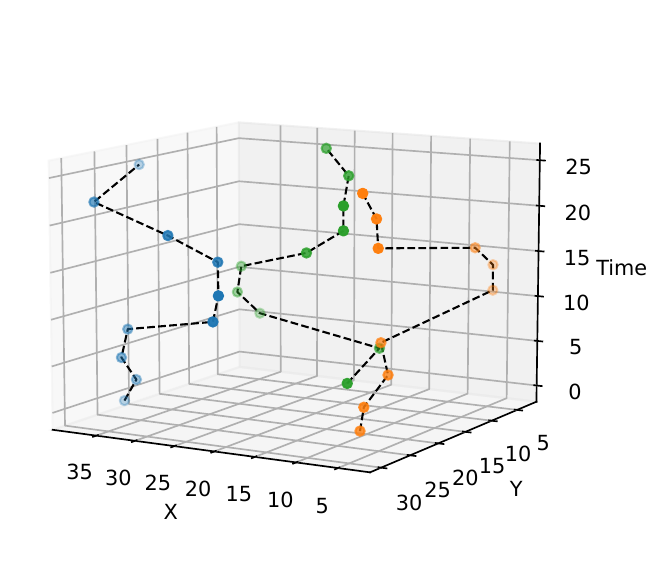}
    \hfill
    \includegraphics[width=0.3\linewidth]{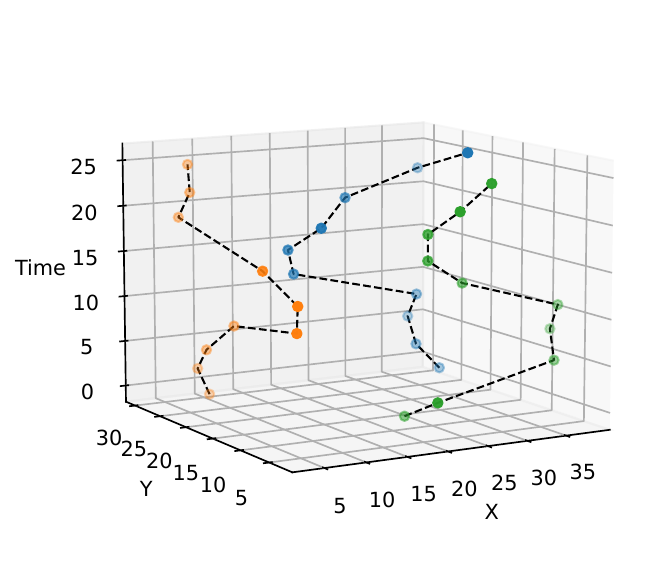}
    \hfill
	\caption{Three different views of our multi-robot IPP solution paths, with path lengths of 47.29 m, 47.44 m, and 47.20 m. The data from all 3 paths gave us an RMSE of 0.34.}
  \label{fig:multi}
\end{figure}

\begin{figure}
    \centering
    \includegraphics[width=0.3\linewidth]{figures/sp_height_4.pdf}
    \hfill
    \includegraphics[width=0.3\linewidth]{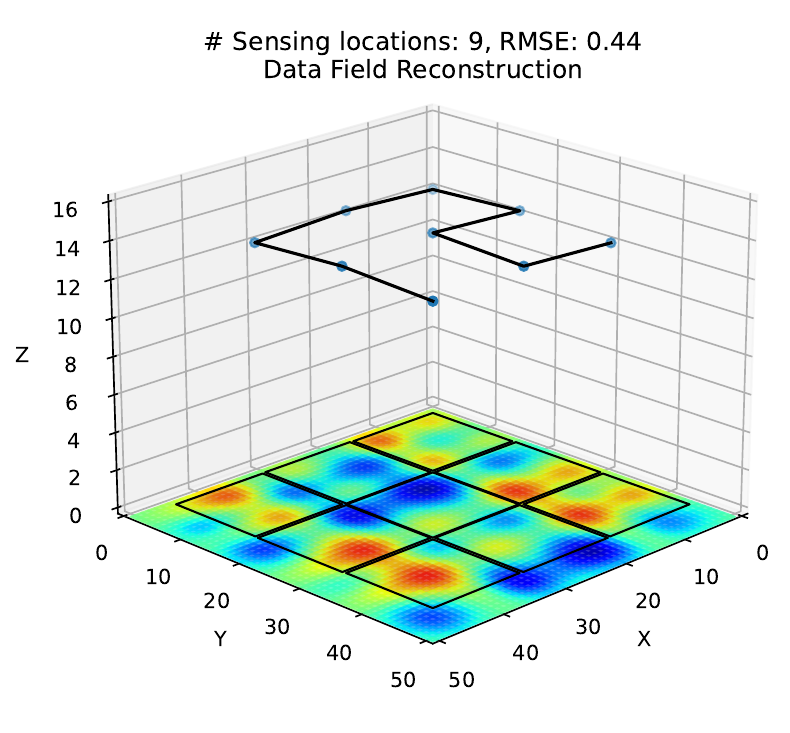}
    \hfill
     \includegraphics[width=0.3\linewidth]{figures/sp_height_36.pdf}
    \hfill
	\caption{Solution paths for a discrete sensing robot with a square height-dependent FoV area (black squares) sensor. The solution paths adjust the sensor height to ensure a good balance between the ground sampling resolution and the coverage area.}
  \label{fig:height}
\end{figure}

\section{Efficient Inference and Parameter Estimation}
Our SGP based IPP approach is only used to obtain informative path(s). Once we have collected data from the path(s), we can use a full GP to estimate the state of the entire environment. However, even if we use a small number of robots for monitoring, the computational cost of using a GP would eventually become infeasible if we are interested in persistent monitoring of an environment. This is due to the cubic complexity of GPs, which is $\mathcal{O}(r^3t^3)$ when considering data from $r$ robots at $t$ timesteps. We can address the cubic complexity issue by using streaming sparse Gaussian process (SSGPs, \cite{BuiNT17}). The approach allows us to efficiently fit a sparse Gaussian process to streaming data sequentially in batches without having to retrain the SGP on the whole dataset. 

\section{Space-Time Decomposition}
When considering multiple robots in spatio-temporal environments, our approach can be further optimized to reduce its computation cost and can be show to have optimal waypoint assignments. When optimizing the paths with $t$ waypoints for $r$ robots, instead of using $rt$ inducing points $\mathbf{X}_m \in \mathbb{R}^{r \times t \times (d+1)}$, we can decouple the spatial and temporal inducing points into two sets—spatial and temporal inducing points. The spatial inducing points $\mathbf{X}_\text{space} \in \mathbb{R}^{rt \times d}$ are defined only in the $d$-spatial dimensions. The temporal inducing points $\mathbf{X}_\text{time} \in \mathbb{R}^{t}$ are defined across the time dimension separately. Also, we assign only one temporal inducing point for the $r$ robots at each time step. This would constrain the $r$ paths to have temporally synchronised waypoints. We then combine the spatial and temporal inducing points by mapping each temporal inducing point to the corresponding $r$ spatial inducing points, forming the spatio-temporal inducing points $\mathbf{X}_m \in \mathbb{R}^{r \times t \times (d+1)}$. 

The approach allows us to optimize the inducing points across space at each timestep and the timesteps separately. It ensures that there are exactly $r$ inducing points at each time step and also reduces the number of variables that need to be optimized. Specifically, we only have to consider $t$ temporal inducing points, rather than $rt$ inducing points along the temporal dimension. During training, we can use backpropagation to calculate the gradients for the decomposed spatial and temporal inducing points through the spatio-temporal inducing points. 

\begin{figure}[ht!]
    \centering
    \includegraphics[width=0.5\linewidth]{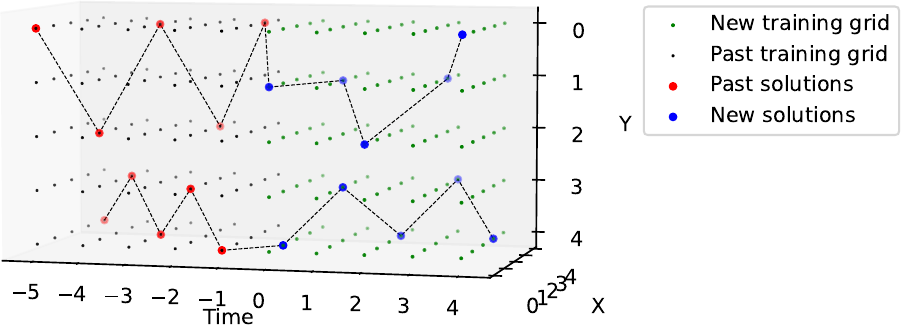}
    \caption{A schematic illustration of the IPP approach for multiple robots (2 robots). Note that the past training grid is only for visual interpretability and is not used while training the SGP to get the new path.}
    \label{fig:multi_path}
\end{figure}

An added advantage of this decomposition is that we can leverage it to prove that the solution paths have optimal waypoint transitions. We do this by setting up an assignment problem that maps the $r$ waypoints at timestep $i$ to the $r$ waypoints at timestep $i+1$. We calculate the assignment costs using pairwise Euclidean distances and get the optimal waypoint transitions for each of the $r$ paths by solving for the assignments~\cite{burkardDM12}. We repeat this procedure for all $t$ timesteps. If the transitions are optimal, the approach would return the original paths, and if not, we would get the optimal solution. The pseudocode of the approach is shown in Algorithm~\ref{alg:mapping}, and Figure~\ref{fig:multi_path} illustrates our approach.

\begin{algorithm}[ht]
\KwIn{Inducing points $\mathbf{X} \in \mathbb{R}^{r \times t \times (d+1)}$}
\KwOut{Waypoints of paths $\mathbf{X} \in \mathbb{R}^{r \times t \times (d+1)}$}
\For{$i\leftarrow 0$ \KwTo $t-1$}{
    $\mathbf{C} = \mathbf{0}^{r \times r}$ \\
    \For{$j\leftarrow 1$ \KwTo $r$}{
        \For{$k\leftarrow 1$ \KwTo $r$}{
             $\mathbf{C}[j][k] \leftarrow  ||\mathbf{X}[j, i, :d]-\mathbf{X}[k, i+1, :d]||_2$
        }
    }
    $A = \mathcal{H}(\mathbf{C})$  \ \text{/ / \tenrm{Solve the assignment problem~\cite{burkardDM12}}} \\
    \text{/ / \tenrm{Re-index the inducing points at time $i+1$}} \\
    $\mathbf{X}[:, i+1, :d+1] \leftarrow \mathbf{X}[A, i+1, :d+1]$ 
}
\Return{$\mathbf{X}$}
\caption{Assignment problem based approach to get optimal waypoint transitions for $r$ paths.}
\label{alg:mapping}
\end{algorithm}


\comment{
\begin{figure*}[ht!]
    \centering
    \includegraphics[width=0.24\textwidth]{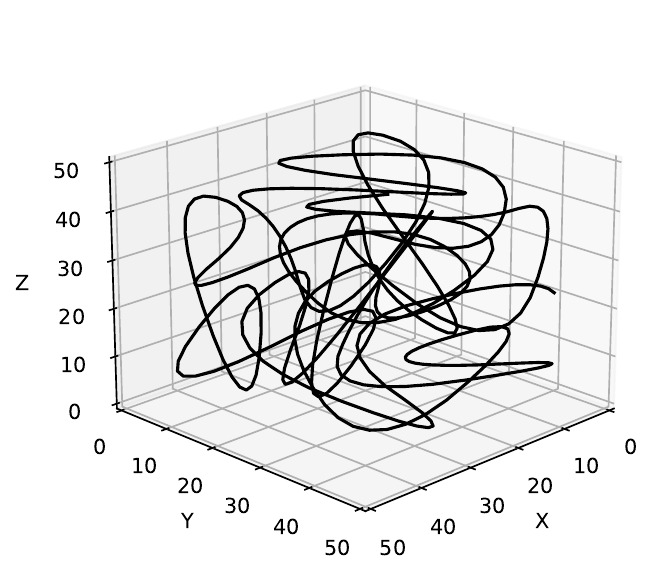}
    \hfill
    \includegraphics[width=0.24\textwidth]{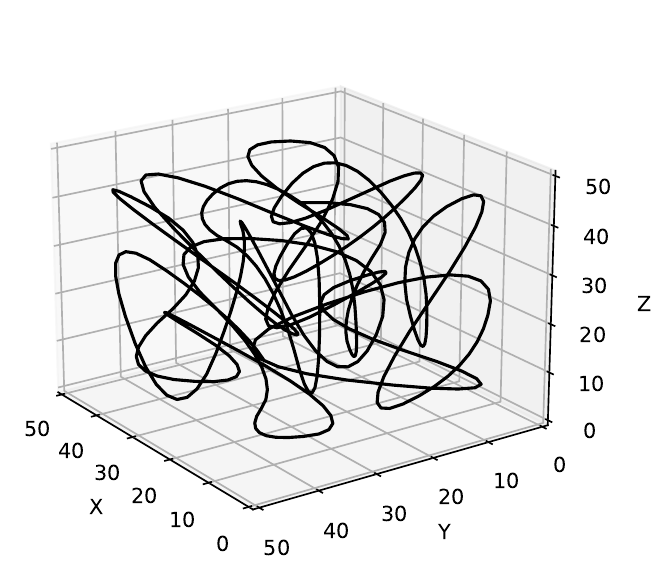}
    \hfill
     \includegraphics[width=0.24\textwidth]{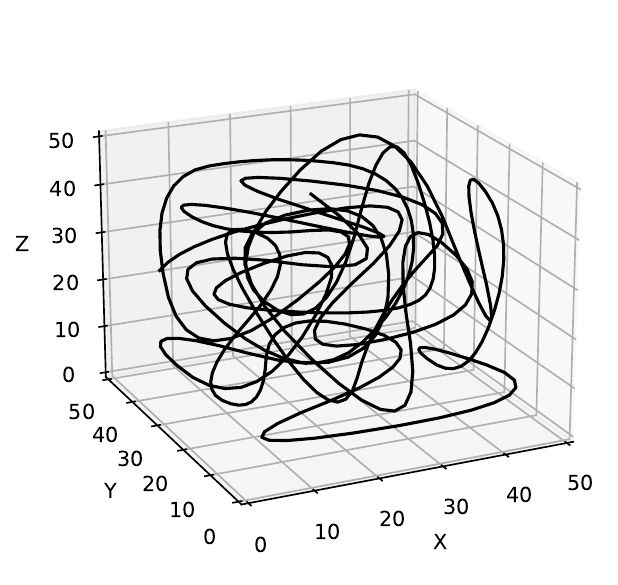}
    \hfill
	\caption{Multiple views of a smooth spline data collection path generated for a continuous sensing robot to cover the whole environment. We see that the path uniformly covers the entire 3D space without being biased towards a sub-region.}
  \label{fig:spline}
\end{figure*}
}

\clearpage
\section{IPP with Past Data}

In a real-world scenario, it is possible that a robot has collected data from a sub-region of the environment. In such cases, in addition to updating our kernel parameters, it would be beneficial to explicitly incorporate the data into our path planning approach. We do this by adding more inducing points—auxiliary inducing points—to the SGP in addition to the $m$ inducing points used to get a path with $m$ sampling locations. The auxiliary inducing points are initialized at the locations of the past data samples, and their temporal dimension is set to negative numbers indicating the time that has passed since the corresponding data sample was collected. During training, the auxiliary inducing points are not optimized; only the original $m$ inducing points used to form the path are optimized. Therefore, the approach accounts for past data, including when it was collected, as the points collected further in the past would be less correlated with the SGP's training data consisting of random samples restricted to the positive timeline. Note that this approach is also suited to address online variants of IPP. 

 \begin{figure*}[ht!]
    \centering
    \includegraphics[width=0.45\textwidth]{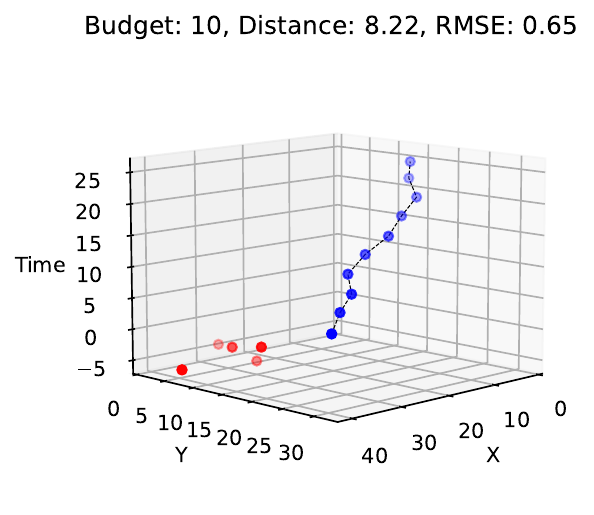}
    \hfill
    \includegraphics[width=0.45\textwidth]{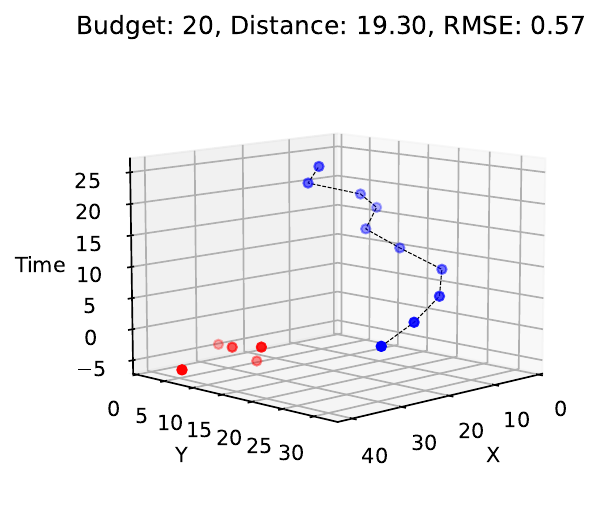}
    \hfill
     \includegraphics[width=0.45\textwidth]{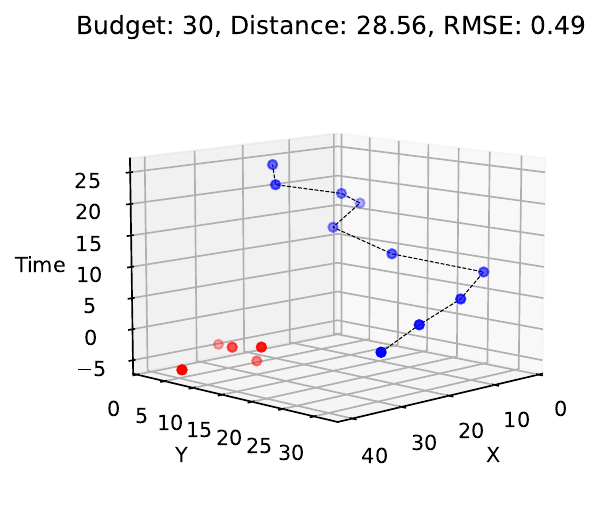}
    \hfill
     \includegraphics[width=0.45\textwidth]{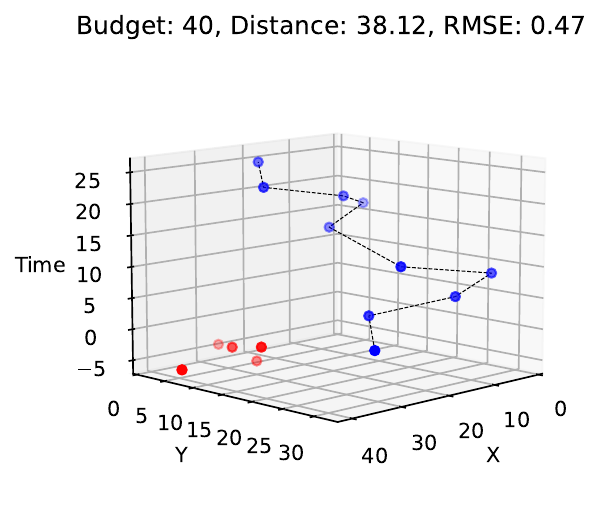}
    \hfill
	\caption{Five points were used as past data (red points). Data collection paths generated using a spatio-temporal kernel function for different distance budgets.}
  \label{fig:hist}
\end{figure*}

We demonstrate the approach with past data. We used 5 data samples as the past data and generated new informative paths with the distance budgets set to 10, 20, 30, and 40. The results are shown in Figure~\ref{fig:hist}; the red points are the past data samples. As we can see, our new solution paths are shifted to avoid collecting data at the location of the past data. Therefore, our solution paths have lower RMSE scores than those generated without past data information.

\clearpage
\section{Linear and Non-linear transformations in SGPs}

This section show how the SGP sensor placement approach~\cite{JakkalaA22b} can be generalized to incorporate the expansion and aggregation transformations to address non-point FoV sensor placement, and in turn, non-point FoV informative path planning. 

\begin{algorithm}[ht]
\setcounter{AlgoLine}{0}
\KwIn{

Hyperparameters $\theta$, domain of the environment $\mathcal{V}$, number of waypoints $s$, number of random unlabeled samples $n$ used to train the SGP, expansion transformation $T_{\text{exp}}$, and aggregation transformation $T_{\text{agg}}$}
\KwOut{Solution sensor placements $\mathcal{A} \subset \mathcal{V}$, where $|\mathcal{A}| = s$}

$\mathbf{X} = \{ \emptyset \}$ \tcp*{\parbox[t]{4.5 in}{\raggedright Initialize empty set to store SGP training set}}
\Repeat{$|\mathbf{X}| = n$}{
 \tcp{Draw $n$ random unlabeled locations from the environment}
$\mathbf{x} \sim \mathcal{U}(\mathcal{V})$ \\
$\mathbf{X} \leftarrow \mathbf{X} \cup \{ \mathbf{x} \}$ \\
}

$\mathcal{D}=(\mathbf{X}, \mathbf{y}=\mathbf{0})$ \tcp*{\parbox[t]{4.5in}{\raggedright Generate SGP training dataset with $0$ labels}}

$\mathbf{X}_m = \text{RandomSubset}(\mathbf{X}, s)$ \tcp*{\parbox[t]{4.5in}{\raggedright Initialize $s$ inducing points at random locations}}
$\mathbf{X}_m  \leftarrow \text{RandomTheta}(\mathbf{X}_m, s)$ \tcp*{\parbox[t]{4.5in}{\raggedright Add random sampled angles as the rotation parameter of each inducing point}}

$\varphi = \mathcal{SGP}(0, \theta, \mathcal{D}, \mathbf{X}_m)$ \tcp*{\parbox[t]{4.5in}{\raggedright Initialize a SVGP $\varphi$ with $0$ mean, hyperparameters $\theta$, training set $\mathcal{D}$, and inducing points $\mathbf{X}_m$}}

\Repeat{convergence}{
$\mathbf{X}_{mp} = T_\text{exp}(\mathbf{X}_{m})$ \tcp*{\parbox[t]{4.5in}{\raggedright Use the expansion transformation $T_{\text{exp}}$ to map the $m$ inducing points $\mathbf{X}_m$ in the point parametrization to $mp$ points with FoV parametrization}}
$\mathbf{Q}_{nn} = (\mathbf{K}_{n \times mp}T_\text{agg}) (T_\text{agg}^\top \mathbf{K}_{mp \times mp}T_\text{agg})^{-1} (T_\text{agg}^\top \mathbf{K}_{mp \times n})$ \tcp*{\parbox[t]{2.6in}{\raggedright Use the aggregation transformation $T_{\text{agg}}$ to reduce the covariances}}
$\mathbf{X}_m \leftarrow \mathbf{X}_m + \gamma \nabla \varphi(\mathbf{Q}_{nn})$ \tcp*{\parbox[t]{4.5in}{\raggedright Optimize the point parametrized inducing points $\mathbf{X}_m$ by maximizing the SVGP's ELBO $\mathcal{F}$ using gradient methods with a learning rate of $\gamma$. We compute the ELBO using the $\mathbf{Q}_{nn}$ computed above}}
}
\Return{$\mathbf{X}_m$}
\caption{Expansion and aggregation transformation based approach for obtaining non-point FoV sensor placements.}
\label{alg:FoV}
\end{algorithm}

Consider a 2-dimensional sensor placement environment. Each of the point parametrized inducing points $\mathbf{X}_m \in \mathbb{R}^{m \times 2}$, are mapped to $p$ points ($\mathbf{X}_mp \in \mathbb{R}^{mp \times 2}$) using the expansion transformation $T_\text{exp}$. This approach scales to any higher dimensional sensor placement environment and can even include additional variables such as the orientation and scale of the sensor/FoV, such as when considering the FoV of a camera on an aerial drone. 

\clearpage
The following is an example of the expansion transformation operation written as a function in Python with TensorFlow. The function considers sensor with a FoV shaped as a line with a fixed length. 

\begin{algorithm}[ht]
\setcounter{AlgoLine}{0}
\SetAlgoLined
\textbf{Input:} $\mathbf{X}_{m}$, $l$, $p$\\
$x, y, \theta = \text{tf.split}(\mathbf{X}_{m}, \text{num\textunderscore or\textunderscore size\textunderscore splits}=3, \text{axis}=1)$ \\
$x = \text{tf.squeeze}(x)$ \\
$y = \text{tf.squeeze}(y)$ \\
$\theta = \text{tf.squeeze}(\theta)$ \\
$\mathbf{X}_{m} = \text{tf.linspace}([x, y], [x + l \times \text{tf.cos}(\theta), y + l \times \text{tf.sin}(\theta)], p, \text{axis}=1))$ \\
$\mathbf{X}_{m} = \text{tf.transpose}(\mathbf{X}_{m}, [2, 1, 0])$ \\
$\mathbf{X}_{m} = \text{tf.reshape}(\mathbf{X}_{m}, [-1, 2])$ \\
\Return{$\mathbf{X}_{m}$}
\caption{Expansion transformation function (written in Python with TensorFlow~\cite{AbadiBCCDDDGII16}) used to map the 2D position ($x, y$) and orientation ($\theta$) to a set of points along a line segment with the origin at the 2D point in the direction of the orientation $\theta$. Here, $\mathbf{X}_m$ are the inducing points with the position and orientation parameterization, $l$ is the length of the line along which the mapped points are sampled, and $p$ is the number of points that are sampled along the line.}
\label{alg:sensor-structure-mask}
\end{algorithm}

The aggregation transformation matrix $T_{\text{agg}} \in \mathbb{R}^{mp \times m}$ is populated as follows for $m = 3$ and $p = 2$ for mean aggregation:

$$
T_{\text{agg}}^\top = \begin{bmatrix}
0.5 & 0.5 & 0 & 0 & 0 & 0 \\ 
0 & 0 & 0.5 & 0.5 & 0 & 0 \\ 
0 & 0 & 0 & 0 & 0.5 & 0.5\\ 
\end{bmatrix} \\. 
$$

However, one can even use the 1-dimensional average pooling operation to efficiently apply the aggregation transformation without having to store large aggregation matrices. 

\bibliography{references.bib}